\renewcommand{\theequation}{\arabic{section}.\arabic{equation}}
\documentclass{ijsipaper}
\if prodtf \else \usepackage{latexsym}\fi
\newcommand\black{\ensuremath{\blacktriangleright}}
\newcommand\white{\ensuremath{\vartriangleright}}
\newif\ifamsfontsloaded
\ifprodtf
  \newcommand\whbl{\white\kern-.1em--\kern-.1em\black}
  \newcommand\blwh{\black\kern-.1em--\kern-.1em\white}
  \newcommand\blbl{\black\kern-.1em--\kern-.1em\black}
  \newcommand\whwh{\white\kern-.1em--\kern-.1em\white}
  \amsfontsloadedtrue
\else
  \checkfont{msam10}
  \iffontfound
    \IfFileExists{amssymb.sty}
      {\usepackage{amssymb}\amsfontsloadedtrue
       \newcommand\whbl{\white\kern-.125em--\kern-.125em\black}%
       \newcommand\blwh{\black\kern-.125em--\kern-.125em\white}%
       \newcommand\blbl{\black\kern-.125em--\kern-.125em\black}%
       \newcommand\whwh{\white\kern-.125em--\kern-.125em\white}}
      {}
  \fi
\fi

\title{Branching Strategy Selection Approach Based on Vivification Ratio }

\author[Mao Luo, Chu-Min Li, Xinyun Wu, Shuolin Li, Zhipeng Lü]
    {Mao Luo${^1}$, Chu-Min Li${^{1,2,4}}$, Xinyun Wu${^3}$, Shuolin Li${^4}$, Zhipeng Lü${^1}$\vspace*{2mm}\\
$^1$ (School of Computer Science, Huazhong Univ. of Science and Technology, \\Wuhan, China, \{maoluo,zhipeng.lv\}@hust.edu.cn)\\
$^2$ (MIS, Univ. of Picardie Jules Verne, Amiens, France, chu-min.li@u-picardie.fr)\\
$^3$ (School of Computer Science, Hubei Univ. of Technology, Wuhan, China, xinyun@hbut.edu.cn)\\
$^4$ (Aix Marseille Univ., Université de Toulon, CNRS, LIS, Marseille, France, li-shuolin@foxmail.com)}

\correspond{Chu-Min Li, Email: chu-min.li@u-picardie.fr}

\pubyear{***} %
\pubmonth{***} %
\pubnumber{***} %
\volume{***} %

\sponsor{the French Agence Nationale de la Recherche, reference ANR-19-CHIA-0013-01} %
\receiveddate{***} %
\reviseddate{***} %
\accepteddate{***} %
\onlinepulishdate{***} %
\paperurl{***} %

\pagerange{\pageref{firstpage}--\pageref{lastpage}}

\usepackage{amssymb}
\usepackage{multirow}
\usepackage{threeparttable}
\usepackage{booktabs}
\usepackage{algpseudocode}
\usepackage[ruled,vlined,linesnumbered]{algorithm2e}
\usepackage{graphicx}
\usepackage{subfigure}
\newtheorem{myDef}{Definition}
\newcommand{\mlcomment}[1]{}

\begin{document}

\setcounter{page}{1}

\label{firstpage}

\maketitle
\makesponsor %
\makecorrespond %
\makeusefuldate %

\begin{abstract}
The two most effective branching strategies LRB and VSIDS perform differently on different types of instances. 
Generally, LRB is more effective on crafted instances, while VSIDS is more effective on application ones. 
However,  distinguishing the types of instances is difficult. 
To overcome this drawback, we propose a branching strategy selection approach based on the vivification ratio. 
This approach uses the LRB branching strategy more to solve the instances with a very low vivification ratio. 
We tested the instances from the main track of SAT competitions in recent years.
The results show that the proposed approach is robust and it significantly increases the number of solved instances. 
It is worth mentioning that, with the help of our approach, the solver Maple\_CM can solve more than 16 instances for the benchmark from the 2020 SAT competition.
\end{abstract}


\makepaperinfo\normalsize\parindent=7mm

\section{Introduction}

The SAT problem consists of finding an assignment to all the variables in a propositional logic formula $\phi$ in conjunctive normal form (CNF), to satisfy all clauses in $\phi$.
SAT is the first problem proven to be NP-complete.
Thus, many NP problems can be solved by transferring them into a SAT equivalent or by considering SAT as a core part of the solving process.
SAT is widely used in various areas, especially in the automation of circuits design, including Equivalence Checking \cite{brand1983redundancy}, Formal Verification \cite{cook2005symbolic}, Automatic Test Pattern Generation \cite{prasad2005survey,drechsler2006automatic}, Model Checking, Logic Synthesis \cite{khomenko2006logic}, software and hardware checking \cite{jackson2000alcoa,clarke2004tool,khurshid2004testera}, planning \cite{kautz1992planning,rintanen2006planning}, and scheduling \cite{lynce2006efficient}.
SAT also affects the research of many related decision and optimization problems \cite{drechsler2006automatic,ganzinger2004dpll,cimatti2008beyond,li2007new,li2010exact}.

For SAT solving, there are two mainstream types of algorithms that are complete algorithms and incomplete algorithms. Complete algorithms could prove unsatisfiability, while incomplete algorithms could not prove unsatisfiability but could solve some certain types of satisfiable instances very efficiently. For complete algorithms, Conflict-Driven Clause Learning (CDCL) solvers and look-ahead solvers achieve success. CDCL solvers could solve industrial application SAT instances very fast; look-ahead solvers are able to solve unsatisfiable random SAT instances efficiently~\cite{heule2016solving,heule2009look,heule2004march_eq}; recently, the combination of CDCL solvers and look-ahead solvers made a breakthrough in automated theory proving~\cite{heule2011cube}. For incomplete algorithms, Stochastic Local Search (SLS) solvers~\cite{cai2014scoring,luo2014double,cai2015ccanr,luo2014ccls,luo2020pbo} and Survey Propagation (SP) solvers are very popular. SLS solvers show strong complementarity with CDCL solvers on solving a number of important application SAT instances, e.g., those instances encoded from station repacking~\cite{cai2021deep,frechette2016solving}; SP solvers show great effectiveness on solving very huge-sized random 3-SAT instances (e.g., with one million variables at the clause-to-variable ratio of 4.2)~\cite{mezard2002analytic,braunstein2005survey,kroc2012survey,gableske2013interpolation,gableske2014ising,gableske2013solver}. This paper focuses on modern CDCL SAT solvers which are very efficient for real applications. 

The core techniques that guarantee the efficiency of the CDCL \cite{marques1999grasp,moskewicz2001chaff} SAT solver include unit propagation, clause learning, branching strategy, clause simplification, management for the database of learnt clauses, restart, and lazy data structure, etc. Clause simplification and branching strategy are among the techniques that gain increasing attention.

The methods of clause simplification can be categorized into pre-processing and in-processing. The most effective pre-processing techniques include variants of Bounded Variable Elimination, Addition or Elimination of Redundant Clauses, Detection of Subsumed Clauses, and suitable combinations of them \cite{een2005effective,bacchus2003effective}.
They aim mostly at reducing the number of clauses, literals, and variables in the input formula.
The most effective in-processing techniques \cite{jarvisalo2012inprocessing} are Local and Recursive Clause Minimization\cite{beame2004towards,sorensson2009minimizing}, On-the-fly Clause Subsumption \cite{han2009fly,hamadi2010learning}, and clause vivification \cite{luo2017effective,li2020clause} where Local and Recursive Clause Minimization removes redundant literals from learnt clauses immediately after their creation, On-the-fly Clause Subsumption efficiently removes clauses subsumed by the resolvents derived during clause learning, and clause vivification periodically detects and removes redundant literals from clauses by unit propagation.

Early branching strategies are based on lookahead and choose the next decision variable by analyzing the clauses not yet satisfied during the searching process.
The most classical branching strategies are MoMs \cite{freeman1995improvements,pretolani1996efficiency}, Dynamic Literal Individual Sum Heuristic(DLIS) \cite{jeroslow1990solving,marques1999impact} and UP \cite{li1997heuristics}. These strategies do not lookback, i.e., they do not learn from what happened in the past to choose the next branching (decision) variable. More recently, lookback branching strategies are introduced in CDCL SAT solvers, consisting in choosing the variables contributing most often to the recent conflicts. The contribution of the variables to the conflicts is collected during the clause learning driven by conflicts. VSIDS (Variable State Independent Decaying Sum)\cite{moskewicz2001chaff} and LRB (Learning Rate Branching)\cite{liang2016learning} are among the best lookback branching strategies.
Different branching strategies may perform quite differently on different categories of instances, and no strategy outperforms others on all the instances.
Many recent CDCL SAT solvers alternatively use VSIDS and LRB to combine their respective strength, independently of the instance to be solved. In these solvers, search is usually divided into pairs of phases, and in each pair of phases, LRB is used to select branching variables in one phase, and VSIDS is used in the other phase. Note that the two phases in the pair have the same length.

We believe that the use of branching strategies should be instance-dependent. In other words, when solving some families of instances, VSIDS should be more often used, while when solving some other families of instances LRB should be more often used. Unfortunately, it is not easy to identify the families of instances for which a particular branching strategy should be used. In this paper, in order to improve the performance of CDCL SAT solvers, we propose to use the information gathered during clause vivification to decide which branching strategy to use more often. The approach is based on the observation that VSIDS appears to outperform LRB when clause vivification discovers many redundant literals in learnt clauses, and LRB appears to outperform VSIDS when clause vivification cannot discover many redundant literals in learnt clauses.

The paper is organized as follows: Section 2 gives some basic concepts about propositional satisfiability and CDCL SAT solvers. Section 3 presents some related works on branching strategies and combination of them. Section 4 gives a detail analysis of the relationship between characteristics of different instances with branching strategy and vivification ratio firstly and then describes our branching strategy selection approach based on vivification ratio, as well as how it is implemented in general CDCL SAT solvers. Section 5 reports on the in-depth empirical investigation of the proposed branching strategy selection approach. Section 6 contains the concluding remarks.

\section{Preliminaries}

 In propositional logic, a variable $x$ may take the truth value 0 (false) or 1 (true). A literal $l$ is a variable $x$ or its negation $\neg x$, a clause $C$ is a disjunction of literals, a CNF formal $\phi$ is a conjunction of clauses, and the size of a clause is the number of literals in it. 
An assignment of truth values to the propositional variables satisfies a literal $x$ if it takes the value 1 and satisfies a literal $\neg x$ if it takes the value 0.
An assignment satisfies a clause if it satisfies at least one of its literals and satisfies a CNF formula if it satisfies all of its
clauses. 
The empty clause, denoted by $\Box$, contains no literal and is unsatisfiable, i.e., it represents a conflict. 
A unit clause contains exactly one literal and is satisfied by assigning the appropriate truth value to the variable. 
An assignment for a CNF formula $\phi$ is complete if each variable in $\phi$ has been assigned a value; otherwise, it is said partial. 
The SAT problem for a CNF formula $\phi$ is to find an assignment to the variables satisfying all clauses of $\phi$.

Modern state-of-the-art SAT solvers are based on the CDCL (Conflict Driven Clause Learning) scheme. The core of the scheme is to continuously generate conflicts and record conflicts through learning clauses. CDCL contains two phases, search phase and learning phase.

In the search phase, the most critical method is the Unit Propagation(UP) method, whose detail is described as follows: 
If there exists a unit clause in $\phi$, to satisfy this clause, the only literal $l$ in it must be satisfied by assigning the appropriate truth value to the corresponding variable.
The satisfaction of  $l$ implies the falsification of $\neg l$.
Therefore, removing the literal $\neg l$ in other clauses and all clauses containing $l$ does not change the satisfiability of the instance.
After the removal, new unit clauses may appear and some clauses can become empty, i.e. no literal remains in the clauses.
We can repeat this process to simplify the instance further until there is no unit clause or an empty clause is produced.
This procedure is denoted as \textsc{UP}$(\phi)$ which returns a simplified formula that does not contain any unit clause, or a formula containing an empty clause.

If the UP procedure does not produce any conflict, the search will heuristically select a variable for assignment to make new UP possible. The selected variable is called \emph{decision variable} and its referred literal is \emph{decision literal}. The level of an assigned variable (by UP or by decision) is the number of decision variables so far. If all the variables are assigned and no conflict occurs after repeating the actions decision and UP, the formula $\phi$ is satisfiable. If an empty clause is produced during the unit propagation, a conflict occurs.

\begin{figure}
	\centering
	\includegraphics[scale=0.3]{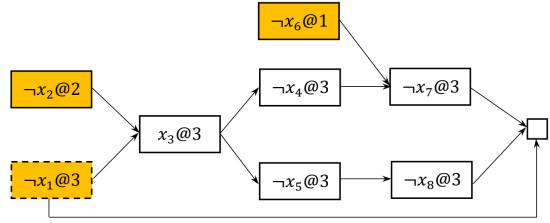}
	\caption{Example of implication graph}
	\label{fig:impGraph}
\end{figure}

The learning phase starts after a conflict occurs. The process of conflict generation can be presented by implication graph as Figure \ref{fig:impGraph}. Each vertex represents the satisfaction of a literal in the figure, marked as $l@dl$, where $l$ denotes the literal, and $dl$ denotes the decision level that the literal belongs to. The negations of the literals of incoming edges of a node $l@ld$ represent the reason (clause) why UP has set $l = 1$. For example, vertex $x_3@3$ is propagated by clause $x_1 \lor x_2 \lor x_3$ ($\neg x_1 \land \neg x_2 \rightarrow x_3$). When the literal $x_1$ and $x_2$ are both assigned  0, $x_3$ must be assigned 1 to satisfy the clause. Conflict clauses are marked as incoming edges to $\Box$, and the level of conflict clauses is named \emph{conflict level}. The decision literal is the vertex marked in orange in the implication graph. It does not have any incoming edge. Note that each decision literal is the starting vertex of its decision level, and the decision literal of the conflict level is dashed. A Unique Implication Point(UIP) is a vertex that is in the conflict level and dominates all the paths to the conflict. Figure \ref{fig:impGraph} contains two UIPs $x_3@3$ and $\neg x_1@3$. Among them, $x_3@3$ is the UIP closest to the conflict, called \emph{first UIP(FUIP)}, $\neg x_1@3$ is the UIP farthest from the conflict, called \emph{last UIP}.

CDCL SAT solvers usually uses the first UIP-scheme for clause learning. Let $dl$ be the conflict level. All literals of level $dl$ on the path from FUIP to the conflict are called \emph{active literals}. For example, in Figure \ref{fig:impGraph}, the conflict level is 3, $\neg x_4@3$, $\neg x_5@3$, $x_7@3$, and $\neg x_8@3$ are active literals. In the FUIP scheme, The negation of each literals with level smaller than $dl$ immediately preceding an active literal, as well as the negation of the FUIP, constructs  the learnt clause. For example, the learnt clause of Figure \ref{fig:impGraph} is $x_6 \lor \neg x_3$. The learnt clause stores the reason for the conflict and allows to avoid repeating the same assignment in the future. After adding the learnt clause, the CDCL solver backtracks to the second highest decision level and performs unit propagation. In Figure \ref{fig:impGraph}, it backtracks to level 1, making the learnt clause $x_6 \lor \neg x_3$ a unit clause $x_6$, and unit clause propagation is performed. If there is a conflict in level 0, it proves that the instance is unsatisfiable.

The variables in the learnt clause $C$ can be partitioned according to its level. The number of partitions in $C$ is called the \emph{Literal Block Distance (LBD)} \cite{audemard2009predicting}. For example, in Figure \ref{fig:impGraph}, the number of levels of the generated learnt clause $x_6@1 \lor \neg x_3@3$ is 2, so its LBD value is 2. A small LBD value usually indicates a clause of good quality.

Let $C=l_1 \vee \cdots \vee l_k$ be a clause in $\phi$, $l_i$ ($1\leq i \leq k$) be a literal in $C$, and $C\setminus l_i$ be $C$ in which $l_i$ is removed, and $\phi'$ be $\phi$ in which $C$ is replaced by $C\setminus l_i$. If $\phi$ and $\phi'$ are equivalent, i.e., any solution of $\phi$ is also a solution of $\phi'$ and vice versa, then $l_i$ is said to be a redundant literal in $C$. Clause vivification is an effective clause simplification technique to remove redundant literals in the clauses of $\phi$, that can be described is as follows.
 If \textsc{UP}$(\phi \cup \{\neg l_1, ..., \neg l_{i-1}, \neg l_{i+1}, ..., \neg l_k\})$ results in an empty clause, then $l_i$ is redundant in $C$.  A clause vivification procedure executing \textsc{UP}$(\phi \cup \neg l_1, \neg l_2, ..., \neg l_i)$ constructs an implication graph, where literals $\neg l_1$, $\neg l_2$, ..., and $\neg l_i$ can be considered as successive decision literals.
For example, if there is a clause $x_1 \lor x_2 \lor x_6 \lor x_9$, such that propagating successively $\neg x_1$, $\neg x_2$ and $\neg x_6$ forms the implication graph as shown in Figure \ref{fig:impGraph}, which implies a conflict. Since the literal $x_9$ does not participate in the conflict, the literal $x_9$ can be deleted from the clause. Learnt Clause Minimization (LCM) approach as presented in \cite{luo2017effective} is a clause vivification technique consisting of eliminating redundant literals from learnt clauses.

\section{Related Works}

Branching is a core process of a CDCL SAT solver (see \cite{freeman1995improvements,pretolani1996efficiency,jeroslow1990solving,marques1999impact,moskewicz2001chaff,ryan2004efficient,goldberg2002and,liang2016exponential,liang2016learning}).
A good branching strategy allows to quickly find the feasible solution of the problem or proves unsatisfactory.
During the solving process of a CDCL SAT solver, the branching strategy analyzes the production of conflicts and guides the search process according to the information gathered during the conflict analysis. 

The most successful branching strategy is Variable State Independent Decaying Sum (VSIDS) \cite{moskewicz2001chaff}.
It computes the score of each variable as follows.
\begin{itemize}
\item The score of each variable is initialized to 0;
\item In a learning phase, the score of each variable encountered in clause learning is increased by $inc$, where $inc$ is a value initialized to 1;
\item After each conflict, $inc$ is increased to $inc/d$, where $d$ is a constant usually fixed to 0.95, so that recent conflicts count more in the score of the variables. 
\end{itemize}

The real applications usually have a clear community structure: the variables within the same community are constrained more strongly than those in different communities \cite{ansotegui2012community}.
Keeping branching on the variables in the same community allows to better analyze the constraint relationship between them. Therefore, the performance of VSIDS on real application instances comes from its capacity to focus on the variables involved in recent conflicts within the same community, because these variables will have high score.

The LRB \cite{liang2016learning} is a new branching strategy based on the Multi-Armed Bandit (MAB) framework in reinforcement learning. The score of each variable is the exponential recency-weighted average of the awards the variable received in the past, as defined below.

\begin{itemize}
	\item The score LRB($v$) of each variable $v$ is initialized to 0;
	\item During the search, a variable is frequently assigned a truth value and this assignment can be canceled later upon backtracking. For variable $v$, let $t_1$ be the number of conflicts produced when $v$ is assigned a truth value, $t_2$ be the number of conflicts produced when this assignment is canceled, and $k$ be the number of conflicts in which $v$ is involved from $t_1$ conflicts to $t_2$ conflicts, then the reward $r$ to $v$ is defined to be $k/(t_2 - t_1)$ and the score LRB($v$) is updated to be $(1-\alpha)*LRB(v) + \alpha*r$, where $\alpha$ is the step-size parameter whose value is empirically initialized to 0.4 and is decreased by $10^{-6}$ after every conflict until it reaches 0.06.
\end{itemize}

Note that LRB considers the $t_2 - t_1$ conflicts in their entirety, while VSIDS emphasizes more on the most recent conflicts because $inc$ for VSIDS is increased after every conflict, which gives LRB stronger global characteristics and VSIDS stronger local characteristics. Because of this difference between VSIDS and LRB, they perform differently on different instances. Currently, no strategy obtains good results on all types of instances. Therefore, before solving an instance, knowing which branching strategy is better for solving it can greatly improve the performance of the solver. Unfortunately, it is hard to know which branching strategy is better for an instance before solving it.
The general useful approach is to use both strategies in the solver.
For example, Maple solver uses LRB for the first half time limit and uses VSIDS for the latter half.
Another solution is to use both strategies alternatively with the phase length multiplied by a growing coefficient.
Apart from the above two approaches, a reinforced learning-based Multi-Armed Bandit (MAB) framework was proposed recently to combine VSIDS and History-based Branching Heuristic (CHB) \cite{liang2016exponential}. Note that CHB is the previous version of LRB without considering learning rate. It selects a proper branching strategy with Upper Confidence Bound (UCB) \cite{cherif2021kissat}. 
The solver using this technique ranked the first place in the main track of SAT Competition 2021.
No matter which technique is used, the combined ones always perform better than the single branching strategy.  
In general, although the technique of using VSIDS and LRB for two half continuous time slots is simple and direct, it performs quite stable for most cases.
And using MAB which self-adaptively chooses the branching strategy according to the conflicts production is also competitive.

\section{Branching Strategy Selection and Learnt Clause Vivification Ratio}
The main purpose of this paper is to explore the relationship between the learnt clause vivification ratio and the two branching heuristics of VSIDS and LRB by analyzing the characteristics of two different types of instances.
The following part first introduces the characteristics of the two types of instances, and then analyzes the relationship between  the vivification ratio of learnt clause and the effectiveness of the branching strategies.
Finally, we introduce the proposed approach in detail based on the above analysis, showing that the branching strategy selecting approach significantly improves the performance of the solver.

\subsection{Two typical classes of SAT instances}
In this section, we describe the two types of instances used in this paper: HWMCC instances generated from real-world EDA applications and PoNo instances crafted from the reduction of the multiplication of two polynomials.

\subsubsection{HWMCC instances}
In various EDA applications, one often needs to check the equivalence of two combinatorial circuits. A common way is to construct a miter circuit from the two original combinatorial circuits such that the unsatisfiability of the miter circuit implies the equivalence of the two original circuits. Since each combinatorial circuit can be formulated as a Direct Acyclic Graph (DAG), in which each internal node of the DAG is one of the seven standard logical operations (AND, OR, NOT, NAND, NOR, XOR, and NXOR), the miter circuit can be encoded into a CNF using the rules of Tseitin transformation\cite{Tseitin1983} defined in Table \ref{tab:encode}.

\begin{table}
\caption{Encode logical operations into CNFs}
\label{tab:encode}
\begin{center}
\begin{threeparttable}
\begin{tabular}{cc}
\toprule
Logical operations & Equivalent CNF Expression\\
\midrule
$C=AND(A,B)$ & $(\neg A \vee \neg B \vee C) \wedge (A \vee \neg C) \wedge (B \vee \neg C)$\\
$C = OR(A,B)$ & $(A \vee B \vee \neg C) \wedge (\neg A \vee C) \wedge (\neg B \vee C)$\\
$C=NOT(A)$ & $(\neg A \vee \neg C) \wedge (A \vee C)$\\
$C = NAND(A,B)$ & $(\neg A \vee \neg B \vee \neg C) \wedge (A \vee C) \wedge (B \vee C)$\\
$C = NOR(A,B)$ & $(A \vee B \vee C) \wedge (\neg A \vee \neg C) \wedge (\neg B \vee \neg C)$\\
$C = XOR(A,B)$ & $(\neg A \vee \neg B \vee \neg C) \wedge (A \vee B \vee \neg C) \wedge (A \vee \neg B \vee C) \wedge (\neg A \vee B \vee C)$\\
$C = NXOR(A,B)$ & $(\neg A \vee \neg B \vee C) \wedge (A \vee B \vee C) \wedge (A \vee \neg B \vee \neg C) \wedge (\neg A \vee B \neg C)$\\
\bottomrule
\end{tabular}
\end{threeparttable}
\end{center}
\label{default}
\end{table}%


The \textit{HWMCC} instances used in this paper are from \textit{SAT Competition 2017} and are provided by Armin Biere (The organizer of the \textit{Hardware Model Checking Competition} \cite{biere2017deep}). 
Armin Biere generates 433 small-scale and 330 large-scale CNF instances from 123 real-world and 12 circuits using AIGUNROLL from the AIGER tools \cite{biere2006aiger}.
By analyzing these instances, they chose 41 challenging ones, which cannot be solved within the time limit by any existing solver before 2017. We use these 41 challenging instances to test the relationships of vivification ratio and different branching strategy between VSIDS and LRB.

\subsubsection{PoNo instances}
For comparison, we introduce the PoNo benchmark which contains 38 SAT instances encoding the problem of multiplying two polynomials of degree $n-1$ with $t$ $(t\leq n^2)$ coefficient products. The initial objective of this encoding is to use SAT solvers to reduce the number of needed coefficient products to multiply two polynomials. And we submitted these instances into 2017 SAT Competition \cite{513c169aa2594baaa57939623626f966}.

A simple example of polynomial multiplication can be expressed by Equation \ref{equ1}:

\begin{equation}\label{equ1}
(ax+b) (cx+d) = acx^{2} + (ad+bc)x+bd
\end{equation}

The trivial multiplication of the two polynomials of degree 1 needs 4 coefficient products: $\{ac, ad, bc, bd\}$. A smart multiplication of the two polynomials needs only 3 coefficient products $\{ac, (a+b)(c+d), bd\}$, as expressed in Equation \ref{equ2}:

 \begin{equation}\label{equ2}
(ax+b) (cx+d) = acx^{2} + \Big( (a+b)(c+d)-ac-bd \Big)x+bd
\end{equation}

In Equation \ref{equ2}, we need more addition and subtraction operations than in Equation \ref{equ1}. However, multiplication is much more costly than addition and subtraction. Thus, we can multiply two polynomials of degree 1 more quickly using Equation \ref{equ2} than using Equation \ref{equ1}.

In the sequel, we describe how to encode the problem of multiplying two polynomials of degree $n-1$ using $t$ ($t\leq n^2$) coefficient products to SAT. When the obtained SAT instance is satisfiable, the SAT solution gives a way to multiply two polynomials of degree $n-1$ using $t$ coefficient products. When the obtained SAT instance is unsatisfiable, we know that more than $t$ coefficient products are needed. We refer to~\cite{BP94,Zippel12} for other efficient algorithms for polynomials.

Consider two polynomials of degree $n-1$:
$$
A(x)= a_{n-1}x^{n-1} + a_{n-2}x^{n-2}+\cdots+a_1x+a_0 
$$
$$
B(x)= b_{n-1}x^{n-1} + b_{n-2}x^{n-2}+\cdots+b_1x+b_0 
$$

Their product is

$$A(x)\times B(x)=c_{2n-2}x^{2n-2} + c_{2n-3}x^{2n-3} + \cdots + c_1x+c_0$$

We want to compute $A(x)\times B(x)$ using $t$ ($t\leq n^2$) coefficient products: $P_1, P_2, \ldots, P_t$, where each $P_l$ ($1\leq l \leq t$) is of the form $(a'_1+a'_2+\cdots)(b'_1+b'_2+\cdots)$ with $a'_1, a'_2, \ldots \in \{a_{n-1}, a_{n-2}, \ldots, a_0\}$ and $b'_1, b'_2, \ldots \in \{b_{n-1}, b_{n-2}, \ldots, b_0\}$. Addition and subtraction of these products give the coefficients $c_k$ ($0\leq k \leq 2n-2$) of $A(x)\times B(x)$. The problem becomes to determinate $a'_i$ and $b'_j$ for each product. In order to solve the problem, we first define the following Boolean variables.  

\begin{itemize}
\item $a_{il}=1$ iff $a_i$ is involved in product $P_l$;
\item $b_{jl}=1$ iff $b_j$ is involved in product $P_l$;
\item $c_{kl}=1$ iff product $P_l$ is used to compute $c_k$;
\item $x_{ijkl}=1$ iff $a_i$ and $b_j$ are involved in product $P_l$, and product $P_l$ is used to compute $c_k$;
\end{itemize}

We then define the clauses, which encode the following properties:

\begin{itemize}
\item $x_{ijkl} \equiv a_{il}\wedge b_{jl} \wedge c_{kl}$
\item For each $i$ and  $j$ ($0\leq i,j \leq n-1)$ and for each $k$ ($0\leq k \leq 2n-2$) such that $i+j \neq k$, if $a_i$ and $b_j$ are involved in product $P_l$ (i.e., $a_{il}\wedge b_{jl}$ is implied) and $P_l$ is used to produce $c_k$, then the product of $a_i$ and $b_j$ should be eliminated by subtraction using another product $P_{l'}$ involving $a_i$ and $b_j$. If $i+j = k$, one product of $a_i$ and $b_j$ should remain in $c_k$. So, 
$$
\sum_{l=1}^t  x_{ijkl} \mbox{ mod } 2 = \left\{
\begin{array}{rl}
0 & \mbox{if } i+j=k\\
1 & \mbox{otherwise}
\end{array}
\right.
$$
\end{itemize}

We generated 38 SAT instances, using the encoding described above, by varying $n$ and $t$. Each combination of $n$ and $t$ gives an instance denoted by N$n$T$t$. The constructed PoNo benchmark differs from the real-world HWMCC benchmark in many ways, which we will describe in Section \ref{sec:observation}.

\subsection{Observation and motivation}\label{sec:observation}
In this section, we present our observations on the differences of HWMMC and PoNo benchmarks with learnt clause vivification ratio.
We analyze the relationship between the vivification ratio and the efficiency of the branching strategies on each type of instances, which constitutes the initial motivation of our proposed approach.

\subsubsection{Redundancies and vivification ratio}




It is widely known that SAT instances often contain redundancies and eliminating these redundancy can greatly help solve the SAT instances. The HWMCC instances contain redundancy in the original CNF formula, because combinatorial circuits can contain redundant logic gates for efficiency, while the PoNo instances do not contain any redundancy in their construction. Nevertheless, when solving an instance, a CDCL solver will add learnt clauses that often contain redundant literals, even for instances that do not contain redundancy initially like the PoNo instances. 



\noindent\textbf{1) Size reduction on HWMCC}

We test Maple\cite{liang2016maple} and Maple\_LCM\cite{xiao2017maplelrb} solvers on the HWMCC instances. Maple solver is based on COMiniSatPs\cite{oh2016cominisatps} which is created by applying a series of small diff patches to MiniSat\cite{een2003extensible}.
The only difference between these two solvers is that Maple\_LCM uses learnt clause vivification named LCM simplification method, while Maple does not.
Note that Maple and Maple\_LCM were the winners of the main track of SAT Competition 2016 and 2017 respectively. 
Table \ref{tab:compare_LRB_LCM_CM} presents the ratio of original redundant literals (denoted as redundant\_ratio) and the vivification ratio (denoted as vivi\_ratio) on learnt clauses by LCM for each instance.

We use the following method to detect redundant literals in each original or learnt clause of each instance.

\begin{enumerate}
\item Set all literals in the clause to be false, except the literal to be checked;
\item Perform unit propagation (UP);
\item If UP produces an empty clause (conflict), the checked literal is redundant in the clause.  
\end{enumerate}

The ratio of redundant literals in the initial CNF formula is computed before starting the instance solving by summing up the number of all detected redundant literals in all original clauses and dividing the obtained sum by the total number of literals in these clauses, and the ratio of redundant literals in learnt clauses is similarly computed among all learnt clauses until the end of the instance solving. These redundant literals are removed from their clauses once detected, performing clause vivification. So the ratio of redundant clauses is also called {\em vivification ratio}.

\begin{table}
\caption{Comparison between Maple and Maple\_LCM on HWMCC}
\label{tab:compare_LRB_LCM_CM}
\scriptsize
\begin{threeparttable}
\begin{tabular}{lrrrr}

\toprule
\multicolumn{1}{c}{\multirow{2}{*}{Instance}} & \multicolumn{2}{c}{CPU Time(s)} & \multicolumn{1}{c}{\multirow{2}{*}{vivi\_ratio}} & \multirow{2}{*}{original redundant\_ratio} \\ \cline{2-3}
\multicolumn{1}{c}{}                          & Maple        & Maple\_LCM       & \multicolumn{1}{c}{}                             &                                   \\
\midrule
6s105-k35.cnf         & 1297.61                       & 964.48                             & 35.85\%                  & 0.26\% \\ 
6s161-k17.cnf         & 4141.58                       & 2749.23                            & 25.45\%                  & 0.43\% \\ 
6s161-k18.cnf         & -                             & -                                  & 23.99\%                  & 0.39\% \\ 
6s179-k17.cnf         & 3066.29                       & 662.4                              & 40.43\%                  & 2.38\% \\ 
6s188-k44.cnf         & 1726.07                       & 985.42                             & 36.76\%                  & 0.58\% \\ 
6s188-k46.cnf         & 2285.93                       & 1012.72                            & 41.58\%                  & 0.55\% \\ 
6s33-k33.cnf          & 1005.14                       & 438.96                             & 43.90\%                  & 0.49\% \\ 
6s33-k34.cnf          & 1019.97                       & 710.65                             & 35.74\%                  & 0.47\% \\ 
6s340rb63-k16.cnf     & 22.58                         & 13.98                              & 34.51\%                  & 3.73\% \\ 
6s340rb63-k22.cnf     & 326.16                        & 199.41                             & 22.35\%                  & 2.90\% \\ 
6s341r-k16.cnf        & 223.36                        & 83.82                              & 32.86\%                  & 1.43\% \\ 
6s341r-k19.cnf        & 1218.09                       & 297.46                             & 28.69\%                  & 1.17\% \\ 
6s366r-k72.cnf        & 2031.58                       & 929.27                             & 39.92\%                  & 0.07\% \\ 
6s399b02-k02.cnf      & -                             & -                                  & 15.06\%                  & 0.02\% \\ 
6s399b03-k02.cnf      & -                             & -                                  & 14.00\%                  & 0.02\% \\ 
6s44-k38.cnf          & 1594.35                       & 772.21                             & 33.54\%                  & 1.54\% \\ 
6s44-k40.cnf          & 2026.48                       & 1286.06                            & 34.89\%                  & 1.51\% \\ 
6s516r-k17.cnf        & 2439.69                       & 1368.93                            & 41.24\%                  & 5.65\% \\ 
6s516r-k18.cnf        & 4967.21                       & 2715                               & 42.83\%                  & 5.43\% \\ 
beembkry8b1-k45.cnf   & 1670.45                       & 725.82                             & 35.35\%                  & 0.14\% \\ 
beemcmbrdg7f2-k32.cnf & -                             & 1961.92                            & 33.93\%                  & 0.32\% \\ 
beemfwt4b1-k48.cnf    & -                             & -                                  & 37.91\%                  & 0.43\% \\ 
beemhanoi4b1-k32.cnf  & -                             & 3244.89                            & 35.19\%                  & 0.16\% \\ 
beemhanoi4b1-k37.cnf  & -                             & -                                  & 29.71\%                  & 0.14\% \\ 
beemlifts3b1-k29.cnf  & 1419.55                       & 957.29                             & 45.08\%                  & 2.80\% \\ 
beemloyd3b1-k31.cnf   & 1992.58                       & 1225.17                            & 35.77\%                  & 1.51\% \\ 
bob12s02-k16.cnf      & -                             & -                                  & 4.07\%                   & 0.00\% \\ 
bob12s02-k17.cnf      & -                             & -                                  & 4.91\%                   & 0.00\% \\ 
bobpcihm-k30.cnf      & -                             & 4907.19                            & 19.04\%                  & 2.39\% \\ 
bobpcihm-k31.cnf      & -                             & -                                  & 19.12\%                  & 2.28\% \\ 
bobpcihm-k32.cnf      & -                             & -                                  & 19.54\%                  & 2.17\% \\ 
bobpcihm-k33.cnf      & -                             & -                                  & 20.27\%                  & 2.07\% \\ 
intel032-k84.cnf      & 4240.67                       & 855.38                             & 31.71\%                  & 0.66\% \\ 
intel065-k11.cnf      & -                             & -                                  & 17.37\%                  & 2.76\% \\ 
intel066-k10.cnf      & -                             & -                                  & 16.04\%                  & 3.02\% \\ 
oski15a10b06s-k24.cnf & -                             & -                                  & 15.58\%                  & 2.44\% \\ 
oski15a10b08s-k23.cnf & -                             & 4395.33                            & 22.70\%                  & 2.79\% \\ 
oski15a10b10s-k20.cnf & -                             & 3055.18                            & 28.02\%                  & 2.75\% \\ 
oski15a10b10s-k22.cnf & -                             & -                                  & 21.12\%                  & 2.58\% \\ 
oski15a14b04s-k16.cnf & 3328.91                       & 368.31                             & 66.03\%                  & 1.20\% \\ 
oski15a14b30s-k24.cnf & 3189.68                       & 1929.16                            & 40.20\%                  & 0.80\% \\ 
\midrule
AVG               & 2056.09                       & 1437.62                            & 29.81\%                  & 1.52\% \\ 
\bottomrule
\end{tabular}
\end{threeparttable}
\end{table}

We observe from Table \ref{tab:compare_LRB_LCM_CM} that most HWMCC instances do not have many original redundant literals (less than 1\%), while the vivification ratio on the learnt clauses is significant (29.78\% on average).
It indicates that the performance of vivification on the learnt clauses is not related very much to the ratio of original redundant literals.
The solver Maple\_LCM (with vivification on learnt clauses) outperforms Maple (without vivification) by solving 5 more instances among the 41 instances.
We also observe that, for the instances containing more than 5\% original redundant literals, e.g., 6s516r-k17 and 6s516r-k18, Maple\_LCM uses much less time compared to Maple.
The reason lies in that the learnt clauses tend to contain redundant literals if they already exist in the original clauses together with the redundant literals resulted in by the learning process, and therefore, the vivification process can simplify the clauses by eliminating these literals.

\noindent\textbf{2) Size reduction on PoNo}

We solve the PoNo instances using both Maple and Maple\_LCM.
The results are shown in Table \ref{tab:pono_red} where the CPU time, the satisfiability, and the vivification ratio for each instance are reported.
The vivification does not help to solve more problem instances,
but it reduces the consumed CPU time for most of the solved instances. 
The average time for Maple is 832.29 seconds but 640.47 seconds for Maple\_LCM.

\begin{table}
\caption{Comparison between Maple and Maple\_LCM on PoNo}
\label{tab:pono_red}
\scriptsize
\centering
\begin{threeparttable}
\begin{tabular}{crrrrr}
\toprule
\multicolumn{1}{c}{\multirow{2}{*}{Instance}} & \multicolumn{2}{c}{Maple} & 
\multicolumn{3}{c}{Maple\_LCM}                 \\ \cline{2-6}

\multicolumn{1}{c}{}                          & CPU Time(s)  & Satisfiability & CPU Time(s) & Satisfiability & vivi\_ratio \\ 
\midrule
N5T06.cnf                                      & 23.98     & UNSAT          & 60.69    & UNSAT          & 21.37\%             \\ 
N5T07.cnf                                      & -         & -              & -        & -              & 9.29\%              \\ 
N5T14.cnf                                      & -         & -              & 1065.81  & SAT            & 4.62\%              \\ 
N5T15.cnf                                      & 930.14    & SAT            & -        & -              & 4.25\%              \\ 
N5T16.cnf                                      & 39.07     & SAT            & 92.15    & SAT            & 3.84\%              \\ 
N6T06.cnf                                      & 37.29     & UNSAT          & 35.86    & UNSAT          & 23.10\%             \\ 
N6T07.cnf                                      & 4646.38   & UNSAT          & 3395.28  & UNSAT          & 14.72\%             \\ 
N6T25.cnf                                      & 995.03    & SAT            & 672.38   & SAT            & 2.63\%              \\ 
N6T27.cnf                                      & 198.78    & SAT            & 99.24    & SAT            & 2.74\%              \\ 
N6T28.cnf                                      & 250.19    & SAT            & 174.44   & SAT            & 2.79\%              \\ 
N6T29.cnf                                      & 1622.49   & SAT            & 165.76   & SAT            & 2.63\%              \\ 
N6T30.cnf                                      & 100.87    & SAT            & 99.94    & SAT            & 2.69\%              \\ 
N7T07.cnf                                      & -         & -              & -        & -              & 10.34\%             \\ 
N7T08.cnf                                      & -         & -              & -        & -              & 7.26\%              \\ 
N7T42.cnf                                      & -         & -              & -        & -              & 1.98\%              \\ 
N7T43.cnf                                      & -         & -              & -        & -              & 1.92\%              \\ 
N7T44.cnf                                      & -         & -              & -        & -              & 1.88\%              \\ 
N7T45.cnf                                      & -         & -              & -        & -              & 1.92\%              \\ 
N7T46.cnf                                      & -         & -              & -        & -              & 1.81\%              \\ 
N8T60.cnf                                      & -         & -              & -        & -              & 1.53\%              \\ 
N8T61.cnf                                      & -         & -              & -        & -              & 1.50\%              \\ 
N8T62.cnf                                      & -         & -              & -        & -              & 1.46\%              \\ 
N8T63.cnf                                      & -         & -              & -        & -              & 1.51\%              \\ 
N11T118.cnf                                    & -         & -              & -        & -              & 1.11\%              \\ 
N13T165.cnf                                    & -         & -              & -        & -              & 0.85\%              \\ 
N13T166.cnf                                    & -         & -              & -        & -              & 0.92\%              \\ 
N14T194.cnf                                    & -         & -              & -        & -              & 0.84\%              \\ 
N27T6.cnf                                      & 153.77    & UNSAT          & 230.33   & UNSAT          & 21.65\%             \\ 
N29T6.cnf                                      & 176.28    & UNSAT          & 207.07   & UNSAT          & 24.73\%             \\ 
N37T6.cnf                                      & 514.25    & UNSAT          & 415.75   & UNSAT          & 22.73\%             \\ 
N39T6.cnf                                      & 618.62    & UNSAT          & 581.28   & UNSAT          & 22.86\%             \\ 
N42T6.cnf                                      & 695.86    & UNSAT          & 555.21   & UNSAT          & 26.05\%             \\ 
N44T6.cnf                                      & 723.79    & UNSAT          & 890.66   & UNSAT          & 21.43\%             \\ 
N45T6.cnf                                      & 654.03    & UNSAT          & 727.18   & UNSAT          & 20.98\%             \\ 
N49T6.cnf                                      & 1095.61   & UNSAT          & 819.04   & UNSAT          & 22.39\%             \\ 
N51T6.cnf                                      & 1379.78   & UNSAT          & 1082.04  & UNSAT          & 21.09\%             \\ 
N52T6.cnf                                      & 1332.57   & UNSAT          & 1036.91  & UNSAT          & 26.48\%             \\ 
N54T6.cnf                                      & 1289.34   & UNSAT          & 1042.88  & UNSAT          & 22.41\%             \\ 
\midrule
AVG                                        & 832.29    & -              & 640.47   & -              & 10.11\%             \\ 
\bottomrule
\end{tabular}
\end{threeparttable}
\end{table}

We can observe that the vivification ratio for the UNSAT instances is higher than for the SAT instances.
Among the UNSAT instances, the ratio is higher for those with a smaller $t$ value. The reason may be that in the UNSAT instances, there are more tight constraints, expecially when $t$ is small.

We also find another characteristic that for the instances with the same $n$ value, they become UNSAT from SAT as the $t$ value increases.
At the same time, the instance size grows significantly.
Here, we test the vivification ratio on the instances with $n=5, t = \{5,6,...,18\}$ of which the results are shown in Table \ref{tab:pono_continuous_red}.  
The results are consistent with the analysis that the vivification ratio decreases from 21\% to 3\% as $t$ increases from 6 to 18.

\begin{table}
\scriptsize
\caption{ The vivification ratio on instances with $n=5, t = \{5,6,...,18\}$}
\label{tab:pono_continuous_red}
\centering
\begin{threeparttable}
\begin{tabular}{crrr}
\toprule
\multirow{2}{*}{Instance} & \multicolumn{3}{c}{Maple\_LCM}                 \\ 
\cline{2-4}
                          & CPU Time(s) & Satisfiability & vivi\_ratio \\ 
\midrule
N5T05.cnf                      & 2.34     & UNSAT          & 16.67\%             \\ 
N5T06.cnf                      & 61.31    & UNSAT          & 21.37\%             \\ 
N5T07.cnf                      & -        & -              & 9.26\%              \\ 
N5T08.cnf                      & -        & -              & 6.47\%              \\ 
N5T09.cnf                      & -        & -              & 6.23\%              \\ 
N5T10.cnf                     & -        & -              & 5.53\%              \\ 
N5T11.cnf                     & -        & -              & 5.27\%              \\ 
N5T12.cnf                     & -        & -              & 4.87\%              \\ 
N5T13.cnf                     & -        & -              & 4.51\%              \\ 
N5T14.cnf                     & 1064.59  & SAT            & 4.62\%              \\ 
N5T15.cnf                     & -        & -              & 4.24\%              \\ 
N5T16.cnf                     & 90.38    & SAT            & 3.84\%              \\ 
N5T17.cnf                     & 81.48    & SAT            & 3.44\%              \\ 
N5T18.cnf                     & 24.89    & SAT            & 3.50\%              \\ 
\bottomrule
\end{tabular}
\end{threeparttable}
\end{table}

\noindent\textbf{3) Conflict index, UP index and vivification ratio}\\

\begin{table}
\scriptsize
\caption{Conflict index, UP index and vivification ratio on HWMCC benchmark}
\label{tab:propagete_depth_HWMCC}
\centering
\begin{threeparttable}

\begin{tabular}{lccc}
\toprule
Instance              & Conflict Index & UP Index & vivi\_ratio \\
\midrule
6s105-k35.cnf         & 6.88           & 52.29           & 35.85\%            \\
6s161-k17.cnf         & 6.65           & 41.61           & 25.45\%            \\
6s161-k18.cnf         & 7.08           & 42.29           & 23.99\%            \\
6s179-k17.cnf         & 12.45          & 20.08           & 40.43\%            \\
6s188-k44.cnf         & 7.82           & 182.82          & 36.76\%            \\
6s188-k46.cnf         & 7.57           & 190.39          & 41.58\%            \\
6s33-k33.cnf          & 10.19          & 12.44           & 43.90\%            \\
6s33-k34.cnf          & 10.74          & 12.46           & 35.74\%            \\
6s340rb63-k16.cnf     & 5.03           & 48.21           & 34.51\%            \\
6s340rb63-k22.cnf     & 5.57           & 88.04           & 22.35\%            \\
6s341r-k16.cnf        & 7.46           & 236.65          & 32.86\%            \\
6s341r-k19.cnf        & 7.03           & 260.78          & 28.69\%            \\
6s366r-k72.cnf        & 8.66           & 619.66          & 39.92\%            \\
6s399b02-k02.cnf      & 9.66           & 9.32            & 15.06\%            \\
6s399b03-k02.cnf      & 8.35           & 9.43            & 14.00\%            \\
6s44-k38.cnf          & 7.38           & 87.38           & 33.54\%            \\
6s44-k40.cnf          & 6.61           & 87.3            & 34.89\%            \\
6s516r-k17.cnf        & 9              & 41.91           & 41.24\%            \\
6s516r-k18.cnf        & 6.57           & 37.03           & 42.83\%            \\
beembkry8b1-k45.cnf   & 22.92          & 10.93           & 35.35\%            \\
beemcmbrdg7f2-k32.cnf & 17.03          & 28.62           & 33.93\%            \\
beemfwt4b1-k48.cnf    & 40.75          & 33.8            & 37.91\%            \\
beemhanoi4b1-k32.cnf  & 13.97          & 53.11           & 35.19\%            \\
beemhanoi4b1-k37.cnf  & 11.26          & 52.78           & 29.71\%            \\
beemlifts3b1-k29.cnf  & 35.16          & 43.5            & 45.08\%            \\
beemloyd3b1-k31.cnf   & 9.68           & 40.24           & 35.77\%            \\
bob12s02-k16.cnf      & 2.33           & 99.6            & 4.07\%             \\
bob12s02-k17.cnf      & 2.27           & 111.62          & 4.91\%             \\
bobpcihm-k30.cnf      & 8.77           & 78.29           & 19.04\%            \\
bobpcihm-k31.cnf      & 7.81           & 84.25           & 19.12\%            \\
bobpcihm-k32.cnf      & 8.52           & 78.83           & 19.54\%            \\
bobpcihm-k33.cnf      & 8.19           & 78.77           & 20.27\%            \\
intel032-k84.cnf      & 7.27           & 30.04           & 31.71\%            \\
intel065-k11.cnf      & 2.97           & 19.21           & 17.37\%            \\
intel066-k10.cnf      & 3.28           & 39.54           & 16.04\%            \\
oski15a10b06s-k24.cnf & 5.11           & 22.83           & 15.58\%            \\
oski15a10b08s-k23.cnf & 4.83           & 22.38           & 22.70\%            \\
oski15a10b10s-k20.cnf & 4.94           & 22.42           & 28.02\%            \\
oski15a10b10s-k22.cnf & 4.85           & 22.43           & 21.12\%            \\
oski15a14b04s-k16.cnf & 10.46          & 152.23          & 66.03\%            \\
oski15a14b30s-k24.cnf & 11.57          & 109.88          & 40.20\%            \\
\midrule
AVG                   & 9.62           & 80.86           & 29.81\%            \\
\bottomrule
\end{tabular}

\end{threeparttable}
\end{table}

\begin{table}
\scriptsize
\caption{Conflict index, UP index and vivification ratio on partial PoNo benchmark}
\label{tab:propagete_depth_PoNo}
\centering
\begin{threeparttable}

\begin{tabular}{lccc}
\toprule
Instance   & Conflict Index & UP Index & vivi\_ratio \\
\midrule
N5T06.cnf & 4.85           & 5.14            & 21.37\%            \\
N5T07.cnf & 4.73           & 4.81            & 9.29\%             \\
N5T14.cnf & 2.75           & 3.76            & 4.62\%             \\
N5T15.cnf & 3.11           & 3.64            & 4.25\%             \\
N5T16.cnf & 3.15           & 3.8             & 3.84\%             \\
N6T06.cnf & 4.88           & 7.44            & 23.10\%            \\
N6T07.cnf & 4.33           & 6.79            & 14.72\%            \\
N6T25.cnf & 2.42           & 1.49            & 2.63\%             \\
N6T27.cnf & 2.41           & 1.49            & 2.74\%             \\
N6T28.cnf & 2.63           & 1.5             & 2.79\%             \\
N6T29.cnf & 2.54           & 1.49            & 2.63\%             \\
N6T30.cnf & 2.5            & 1.49            & 2.69\%             \\
N7T07.cnf & 4.27           & 9.73            & 10.34\%            \\
N7T08.cnf & 3.82           & 9.65            & 7.26\%             \\
N7T42.cnf & 2.22           & 1.5             & 1.98\%             \\
N7T43.cnf & 2.35           & 1.5             & 1.92\%             \\
N7T44.cnf & 2.44           & 1.5             & 1.88\%             \\
N7T45.cnf & 2.47           & 1.5             & 1.92\%             \\
N7T46.cnf & 2.21           & 1.49            & 1.81\%             \\
\midrule
AVG        & 3.16           & 3.66            & 6.40\%             \\
\bottomrule
\end{tabular}

\end{threeparttable}
\end{table}

We observe that, during the search process, the implication graphs of HWMCC instances are much more complex than those of PoNo ones.
So, we define two indicators to reflect the differences in the characteristics of these two instances.

\begin{myDef}
\textbf{Conflict Index}
It is the average number of conflict level literals involved in a conflict, that is, the number of active literals. 
\end{myDef}

\begin{myDef}
\textbf{UP Index}
It is the average number of literals propagated by UP after branching on a decision literal.
\end{myDef}


Table \ref{tab:propagete_depth_HWMCC} and Table \ref{tab:propagete_depth_PoNo} show the relationship between the conflict index, UP index and vivification ratio of the HWMCC and PoNo instances, respectively. 
We observe that the average UP index on HWMCC instances is 80.86, while for PoNo instances the average value is 3.66.
The probability of producing conflicts on HWMCC instances is much higher than that on PoNo since the average UP index on HWMCC instances is one order of magnitude larger than that on PoNo.

Compared with the UP index, the conflict index and the vivification ratio are even more consistent. For example, for the instances bob12s02-k16.cnf and bob12s02-k17.cnf, the conflict indexes respectively are 2.33 and 2.27, which are much lower than those of other HWMCC instances. The vivification ratios of these two instances are 4.07\% and 4.91\%, which are also far lower than those of other HWMCC instances. In addition, for the instances oski15a14b04s-k16.cnf and oski15a14b30s-k24.cnf, the conflict indexes are 10.46 and 11.57 respectively. The vivification ratios of these two instances are 66.03\% and 40.2\%, which are also greater than those of other HWMCC instances.

\subsubsection{Branching strategies and vivification ratio}

\mlcomment{
The LBD value of learnt clauses approximately reflects the community structure of the instance.
The LBD value of one clause represents the number of levels in that clause.
We consider that the literals in one community are in the same layer since they are closely connected.
The LBD value of a learnt clause estimates how many communities it connects.
For example, a clause is considered connected to 3 communities if its LBD value is 3.
}

The LBD of a learnt clause is roughly the number of decisions needed to produce a conflict by UP.
Table \ref{tab:LBD} shows the LBD and size of the learnt clauses when solving the HWMCC instances and the PoNo instances using VSIDS and LRB strategies, respectively. It also presents the number of solved instances by Maple\_LCM only uses one branching strategy between VSIDS (denoted by Maple\_LCM/VSIDS) and LRB (denoted by Maple\_LCM/LRB). Note that Maple\_LCM uses LRB for the first half time and VSIDS for the second half time.

\begin{table}
\scriptsize
\caption{LBD of the learnt clauses of two benchmarks using VSIDS and LRB strategies}
\label{tab:LBD}
\centering
\begin{threeparttable}
\begin{tabular}{lcccc}
	\toprule
	Instances                 & \multicolumn{2}{c}{HWMCC}                   & \multicolumn{2}{c}{PoNo}                    \\
	\midrule
	Solvers                   & Maple\_LCM/VSIDS     & Maple\_LCM/LRB       & Maple\_LCM/VSIDS     & Maple\_LCM/LRB       \\
	\midrule
	learnts\_LBD         & 9.59                 & 14.39                & 25.06                & 56.81                \\
	learnts\_Size        & 31.74                & 52.14                & 37.45                & 80.49                \\
	vivi\_LBD            & 3.93                 & 3.98                 & 4.31                 & 4.42                 \\
	vivi\_Size   & 16.85                & 15.82                & 9.37                 & 8.53                 \\
	vivi\_ratio & 34.91\%              & 28.92\%              & 7.46\%               & 7.44\%               \\
	\midrule
	nbSloved                  & \textbf{32}                   & 19                   & 13                   & \textbf{21}                   \\
	\bottomrule
				  & \multicolumn{1}{l}{} & \multicolumn{1}{l}{} & \multicolumn{1}{l}{} & \multicolumn{1}{l}{} \\
				  & \multicolumn{1}{l}{} & \multicolumn{1}{l}{} & \multicolumn{1}{l}{} & \multicolumn{1}{l}{} \\
				  & \multicolumn{1}{l}{} & \multicolumn{1}{l}{} & \multicolumn{1}{l}{} & \multicolumn{1}{l}{} \\
				  & \multicolumn{1}{l}{} & \multicolumn{1}{l}{} & \multicolumn{1}{l}{} & \multicolumn{1}{l}{} \\
				  & \multicolumn{1}{l}{} & \multicolumn{1}{l}{} & \multicolumn{1}{l}{} & \multicolumn{1}{l}{}\\

\end{tabular}
\end{threeparttable}
\end{table}

\begin{table}
\scriptsize
\caption{Results of crafted instances using VSIDS and LRB strategies}
\label{crafted}
\centering
\begin{tabular}{lrrlrr}
\toprule
\multirow{2}{*}{Instance}       & \multicolumn{2}{c}{Maple\_LCM/VSIDS}                              & \multirow{2}{*}{} & \multicolumn{2}{c}{Maple\_LCM/LRB}                                \\ \cline{2-3} \cline{5-6} 
                                & \multicolumn{1}{l}{CPU Time(s)} & \multicolumn{1}{l}{vivi\_ratio} &                   & \multicolumn{1}{l}{CPU Time(s)} & \multicolumn{1}{l}{vivi\_ratio} \\
                                \midrule
crafted\_n10\_d6\_c3\_num18.cnf & -                               & 34.17\%                         & \multirow{3}{*}{} & 331.26                          & 31.25\%                         \\
crafted\_n11\_d6\_c4\_num19.cnf & 976.18                          & 38.60\%                         &                   & 203.43                          & 36.20\%                         \\
crafted\_n12\_d6\_c4\_num9.cnf  & 542.17                          & 45.92\%                         &                   & 226.82                          & 41.34\%                         \\ 
\bottomrule
\end{tabular}
\end{table}

We can make the following two observations from Table \ref{tab:LBD}.

\begin{enumerate}
\item The LBD of the learnt clauses (denoted as learnts\_LBD) in the HWMCC instances is much smaller than that in PoNo, i.e, the number of decisions needed to produce a conflict is much smaller when solving the HWMCC instances than when solving the PoNo instances, because the UP index and the conflict index of the HWMCC instances is much greater than those of the PoNo instances. In fact,
the average LBD of HWMCC instances from the Maple\_LCM/VSIDS solver is 9.59, while this value for PoNo is 25.06. When vivifing a learnt clause $\l_1 \lor \l_2 \lor \cdots \lor l_k$ by successively propagating $\neg l_1, \neg l_2, \ldots, \neg l_i (i \textless k)$, these negated literals can be considered as decisions, and fewer negative literals are needed to be propagated to produce a conflict in the HWMCC instances than in the PoNo instances. 
For the vivified learnt clauses which are selected by LCM approach from the learnt clause database with smaller LBD value, we can see that the size of vivified learnt clauses (denoted as vivi\_Size) in the HWMCC instances is greater than the one in the PoNo instances, on the other hand, vivified learnt clauses in HWMCC and PoNo are roughly have the same LBD value (denoted as vivi\_LBD). Note that all the above values are average. So, simplifying the learnt clauses is usually much easier for the HWMCC instances than for the PoNo instances, explaining the higher vivification ratio for the HWMCC instances. 
\item The VSIDS branching strategy is much better than the LRB branching strategy for the HWMCC instances, while the LRB branching strategy is much better than the VSIDS one for the PoNo instances.
\end{enumerate}

The above two observations suggest that VSIDS should be more used for instances on which the learnt clause vivification ratio is great, and LRB should be more used for instances on which the learnt clause vivification ratio is small. However, some instances with high vivification ratio may also be crafted instances. For example, the 3 instances which are crafted\_n10\_d6\_c3\_num18.cnf, crafted\_n11\_d6\_c4\_num19.cnf and crafted\_n12\_d6\_c4\_num9.cnf from main track of 2020 SAT competition have very high vivification ratios. We test these instances with the two solvers Maple\_LCM/VSDIS and Maple\_LCM/LRB. It can been seen that the performance of only using LRB branching strategy is much better than only using VSIDS one (as shown in Table~\ref{crafted}).

These instances which are typical crafted ones are encoded through representing the graph isomorphism of two graphs. The original problems corresponding to these instances are obvious UNSAT problems. And when they are encoded into SAT instances, the redundancy and the vivification ratio are significantly high in searching process. This phenomenon can also be seen in PoNo instance N27T6.cnf. Although the vivification ratio of N27T6.cnf is more than 20\%, it is still a crafted one. Therefore, in this paper we consider to use more LRB branching strategy for the instances with low vivification ratio. We will propose an approach in this direction in the next section.

\subsection{Branching strategy selection based on the vivification ratio}

The searching procedure of the CDCL SAT solver is shown in Algorithm~\ref{CDCL}.
The SAT solver performs the preprocessing of the instance firstly, including deleting variables, clauses and literals.
Then during  search, the learnt clause vivification will be performed at certain beginning of restarts. 
In the main searching procedure, unit propagation is performed firstly. 
If conflicts occur, conflict analysis is triggered, a new learnt clause is generated, based on which the solver backtracks. 
If there is no conflict, the solver selects a suitable variable to assign a truth value using a branching strategy. 
This process is repeated until a conflict is produced in the 0th level to prove the unsatisfiability, or an assignment satisfying all clauses.

In this section, we describe our approach to allow Algorithm~\ref{CDCL} to choose a suitable branching strategy based on vivification ratio during search, by calling function chooseBranch$(viviRatio)$ (described in Algorithm~\ref{chooseBranch}).

\begin{algorithm}[!h]
\KwIn{$\phi$: A CNF formula with original and learnt clauses.}
\KwOut{SATISFIABLE or UNSATISFIABLE}
\Begin{
	$\phi \leftarrow$ preprocessing$(\phi)$:\\
	\While {true} {
	    $S \leftarrow$ chooseBranch$(viviRatio)$; /* call Algorithm~\ref{chooseBranch} to choose a branching strategy*/\\
		$currentLevel \leftarrow 0$; /* start or restart search */\\
		$\phi \leftarrow$ vivification$(\phi)$; /* select some learnt clauses to vivify */\\
		\While {true} {
			$cl \leftarrow$ UP$(\phi)$; /* all variables assigned by UP are recorded with $currentLevel$ */\\
			\If {$cl$ is a falsified clause}{
				\If {$currentLevel == 0$}{
					return UNSATISFIABLE;
				}
				\Else{
					$newLearntClause \leftarrow$ analyze$(cl)$; /* conflict analysis to learn a new clause */\\
					$level \leftarrow$ the second highest level in $newLearntClause$;\\
					backtrackTo$(level)$; /* cancel all variable assignments higher than level $level$ */\\
					$currentLevel \leftarrow level$;
				}
			}
			\Else{
				\If {all variables are assigned}{
					return SATISFIABLE;\\
				}
				\ElseIf{restart condition is satisfied} {
					backtrackTo(0); /* cancel all assignments depending on a decision */\\
					break; /* restart */
				}
				\ElseIf {learnt clause database reduction condition is satisfied} {
					remove a subset of learnt clauses;
				} 
				\Else{
				    $currentLevel++$;\\
					$x \leftarrow$ a non-assigned variable selected according to the strategy $S$;\\
					add the unit clause $x$ or $\neg x$ into $\phi$ according to a polarity heuristic such as phase saving;
				}
			}
		}
	}
}
\caption{CDCL$(\phi)$, a generic CDCL SAT algorithm \label{CDCL}}
\end{algorithm}

\begin{algorithm}[!h]
\KwIn{viviRatio: vivification ratio of learnt clauses.}
\KwOut{LRB or VSIDS branching strategy}
\Begin{
	\If {the number of restarts since the last execution of this function exceeds $\gamma$}{
		\If {viviRatio $\textless$ $\alpha$}{
			$P_b \leftarrow \beta$; /* $\alpha = 8\%$ and $\beta = 0.8$ */\\
		}
		\Else{
			$P_b \leftarrow 0.5$;
		}
	    $\gamma \leftarrow \gamma + \gamma / 10$;\\
	    with probability $P_b$, return the LRB strategy; else return the VSIDS strategy;\\
	}
}
\caption{chooseBranch$(viviRatio)$, choose a branching strategy \label{chooseBranch}}
\end{algorithm}

Concretely, the purpose of the chooseBranch$(viviRatio)$ function is to determine a probability $P_b$ to select the LRB or VSIDS branching strategy, depending on the vivification ratio of learnt clauses in the last $\gamma$ restarts since the last execution of the chooseBranch$(viviRatio)$ function or from the beginning, where $\gamma$ is a parameter initialized to 10000 and is increased by 10\% each time $P_b$ is re-evaluated. 
 When the learnt clause vivification ratio is lower than $\alpha$, where $\alpha$ is a parameter fixed to 8\%, then $P_b$ is set to 0.8 to select the LRB branching strategy. Otherwise, $P_b$ is set to 0.5. Finally Algorithm 2 returns the LRB strategy with probability $P_b$ and the VSIDS strategy with probability $1-P_b$.

\section{Experiments}
In this section, we evaluate our proposed approach on several SAT Competition benchmarks.

\subsection{Experimental protocol}
The test suite includes the instances from main track (application + crafted) of the SAT Competition 2017, 2018 and 2020. 
The experiments were performed on Intel Xeon E5-2680 v4 processors with 2.40 GHz and 20 GB of memory under Linux. 
The cutoff time is 5000 seconds for each solver and each instance, including the preprocessing time and the search time, unless stated.
For each solver and each benchmark, we report the number of solved SAT/UNSAT instances and total solved instances, denoted as ‘\#SAT’, ‘\#UNSAT’ and ‘\#Solved’, and the penalized run time ‘PAR2’ (as used in SAT Competitions), where the run time of a failed run is penalized as twice the cutoff time. 

The tested approach is implemented based on SAT solver Maple\_CM which is an improved version of Maple\_LCM solver. The difference from Maple\_LCM is that Maple\_CM solver uses vivification for more in-depth clause simplification. In fact, Maple\_CM not only simplifies learnt clauses more than one time using vivification, but also simplifies original clauses during pre-processing and in-processing. Note that the performance of Maple\_CM is greatly improved compared to Maple\_LCM, and it won the third place in the main track of 2018 SAT competition.

\subsection{Efficiency analysis}

In this part, we implemented Algorithm \ref{chooseBranch} on top of the solver Maple\_CM. The resulting solver is named Maple\_CM+. Besides, we created the solver Maple\_CM+/$P_b0.5$ by setting the parameter $P_b$ to the fixed value 0.5 (set $\beta$ as 0.5 in Algorithm \ref{chooseBranch}). In other words, Maple\_CM+/$P_b0.5$ is Maple\_CM+ except that it selects LRB and VSIDS with the same probability. 

\begin{table}
\scriptsize
\caption{ Comparison of the branching strategy selection on different reduction ratios}
\label{tab:reduction ratio comparison}
\centering
\begin{threeparttable}
\begin{tabular}{clllll}
	\toprule
	Instances                                                               & Solver                                & \#Total & \#SAT & \#UNSAT & PAR2 (s) \\ 
	\midrule
	\multirow{3}{*}{\begin{tabular}[c]{@{}c@{}}SC2020\\ (400)\end{tabular}} & Maple\_CM                             & 196     & 90    & 106     & 5672.11  \\
										& Maple\_CM+/$P_b$0.5		         	& 219     & 108   & 111     & 5022.82  \\
										& Maple\_CM+ 				& \textbf{224}     & \textbf{111}   & \textbf{113}     & \textbf{5006.07}  \\ 
	\midrule
	\multirow{3}{*}{\begin{tabular}[c]{@{}c@{}}SC2018\\ (400)\end{tabular}} & Maple\_CM                             & 229     & 129   & 100     & 4643.99  \\
										& Maple\_CM+/$P_b0.5$          			& 233     & 133   & 100     & 4653.62  \\
										& Maple\_CM+ 				& \textbf{237}     & \textbf{136}   & \textbf{101}     & \textbf{4536.32}  \\ 
										\midrule
	\multirow{3}{*}{\begin{tabular}[c]{@{}c@{}}SC2017\\ (350)\end{tabular}} & Maple\_CM                             & \textbf{227}     & \textbf{110}   & \textbf{117}     & \textbf{4116.99}  \\
										& Maple\_CM+/$P_b0.5$          			& 218     & 101   & 117     & 4313.60  \\
										& Maple\_CM+ 				& 224     & 108   & 116     & 4163.75  \\ 
										\bottomrule
	\end{tabular}
\end{threeparttable}
\end{table}

Table~\ref{tab:reduction ratio comparison} compares the solvers Maple\_CM, Maple\_CM+/$P_b0.5$ and Maple\_CM+ on instances of main track of SAT Competition (noted as SC) 2017, 2018 and 2020. The proposed Branching Strategy Selection (BSS) approach obviously improves the performances of all solvers on instances of both SC2018 and SC2020. In particular, for instances of SC2020, Maple\_CM+ can solve more than 18 instances compared with Maple\_CM and the PAR2 solution time is also greatly reduced with the help of the BSS approach. On the other hand, we can see that Maple\_CM+ performs consistently better than Maple\_CM+/$P_b0.5$ for all instances of every SAT competition. 

However, the performance of Maple\_CM+ is worse than that of Maple\_CM for the SC2017 instances. The main reason lays on that Maple\_CM uses LRB for the first 2500 seconds and VSIDS for the last 2500 seconds. While Maple\_CM+ selects LRB or VSIDS with a certain probability after a period of conflict. And there are many application instances of SC2017, which are more suitable for long-term use of VSIDS branching strategy to solve. Therefore, the solving results of Maple\_CM  are better compared to Maple\_CM+ in this special case. On the other hand, for the SC2017 instance, Maple\_CM+ can solve 6 more instances than Maple\_CM+/$P_b$0.5, which shows that the method proposed in this paper is still effective in terms of the choice of branching strategies.

\subsection{Robustness analysis}


Here we tested Maple\_CM+ by fixing $P_b$ from 0\% to 100\% at a growth rate of 10\% (i.e. set $\beta$ from 0\% to 100\% in Algorithm \ref{chooseBranch}). Note that $P_b=0\%$ means that Maple\_CM+ only uses the VSIDS branching strategy, while $P_b=100\%$ means that Maple\_CM+ only uses the LRB branching strategy. For every value of $P_b$, we ran Maple\_CM+ to solve the instances with low vivification ratio ($vivi\_ratio<8\%$) in SC2020, SC2018 and SC2017. Note that there are 134, 189 and 164 instances with low vivification ratio in SC2020, SC2018 and SC2017, respectively. 

\begin{table}
\scriptsize
\caption{Robustness analysis}
\label{tab:robustness analysis}
\centering
\begin{threeparttable}
\begin{tabular}{clllll}
	\toprule
	Instances                                                                          & Solver                                & \#Total & \#SAT & \#UNSAT & PAR2(s) \\ 
	\midrule
	\multirow{11}{*}{\begin{tabular}[c]{@{}c@{}}SC2020\\ (vivi\_low 134)\end{tabular}} & Maple\_CM+/$P_b$0                          & 41      & 16    & 25      & 7450.86 \\
											   & Maple\_CM+/$P_b$0.1 & 49      & 24    & 25      & 6932.79 \\
											   & Maple\_CM+/$P_b$0.2 & 51      & 26    & 25      & 6784.91 \\
											   & Maple\_CM+/$P_b$0.3 & 50      & 23    & 27      & 6806.67 \\
											   & Maple\_CM+/$P_b$0.4 & 49      & 24    & 25      & 6860.51 \\
											   & Maple\_CM+/$P_b$0.5 & 54      & 27    & 27      & 6463.03 \\
											   & Maple\_CM+/$P_b$0.6 & 55      & 29    & 26      & 6397.19 \\
											   & Maple\_CM+/$P_b$0.7 & 58      & 31    & 27      & 6296.81 \\
											   & Maple\_CM+/$P_b$0.8 & \textbf{58}      & \textbf{32}    & 26      & \textbf{6251.66} \\
											   & Maple\_CM+/$P_b$0.9 & 58      & 31    & 27      & 6284.46 \\
											   & Maple\_CM+/$P_b$1                         & 57      & 29    & \textbf{28}      & 6266.66 \\ 
											   \midrule
	\multirow{11}{*}{\begin{tabular}[c]{@{}c@{}}SC2018\\ (vivi\_low 189)\end{tabular}} & Maple\_CM+/$P_b$0                          & 46      & 32    & 14      & 7801.30 \\
											   & Maple\_CM+/$P_b$0.1 & 59      & 45    & 14      & 7168.91 \\
											   & Maple\_CM+/$P_b$0.2 & 65      & 50    & 15      & 6883.54 \\
											   & Maple\_CM+/$P_b$0.3 & 74      & 58    & 16      & 6511.65 \\
											   & Maple\_CM+/$P_b$0.4 & 70      & 55    & 15      & 6541.22 \\
											   & Maple\_CM+/$P_b$0.5 & 70      & 55    & 15      & 6540.55 \\
											   & Maple\_CM+/$P_b$0.6 & 72      & 56    & 16      & 6494.40 \\
											   & Maple\_CM+/$P_b$0.7 & 78      & 62    & 16      & 6213.59 \\
											   & Maple\_CM+/$P_b$0.8 & 75      & 58    & 17      & 6382.57 \\
											   & Maple\_CM+/$P_b$0.9 & 73      & 56    & \textbf{17}      & 6393.49 \\
											   & Maple\_CM+/$P_b$1              & \textbf{81}      & \textbf{65}    & 16      & \textbf{6102.91} \\
											   \midrule
	\multirow{11}{*}{\begin{tabular}[c]{@{}c@{}}SC2017\\ (vivi\_low 164)\end{tabular}} & Maple\_CM+/$P_b$0                          & 46      & 28    & 18      & 7590.08 \\
											   & Maple\_CM+/$P_b$0.1 & 57      & 38    & 19      & 6976.40  \\
											   & Maple\_CM+/$P_b$0.2 & 57      & 38    & 19      & 6894.34 \\
											   & Maple\_CM+/$P_b$0.3 & 53      & 34    & 19      & 7072.37 \\
											   & Maple\_CM+/$P_b$0.4 & 56      & 37    & 19      & 6951.51 \\
											   & Maple\_CM+/$P_b$0.5 & 62      & 42    & 20      & 6678.19 \\
											   & Maple\_CM+/$P_b$0.6 & 61      & 41    & 20      & 6686.56 \\
											   & Maple\_CM+/$P_b$0.7 & \textbf{73}      & \textbf{52}    & 21      & \textbf{6189.76} \\
											   & Maple\_CM+/$P_b$0.8 & 68      & 46    & 22      & 6366.73 \\
											   & Maple\_CM+/$P_b$0.9 & 69      & 46    & \textbf{23}      & 6316.76 \\
											   & Maple\_CM+/$P_b$1                         & 62      & 40    & 22      & 6652.67 \\
											   \bottomrule
\end{tabular}
\end{threeparttable}
\end{table}

Table \ref{tab:robustness analysis} shows that the performance of the solver only using LRB branching strategy is substantially better than the performance of the solver only using VSIDS. And with the fixed value of $P_b$ increases gradually, the result of Maple\_CM+/$P_b$ is also continuously improved. When $P_b$ set as 0.7 or 0.8, the result roughly reaches the highest point. The results show that LRB should be more used for instances on which the learnt clause vivification ratio is small.

We can also see from Table~\ref{tab:reduction ratio comparison} and Table~\ref{tab:robustness analysis} that, in general, BSS approach which is implemented in CDCL SAT solvers can significantly improve the results of SAT instances, while the results of UNSAT instances are not improved obviously. The main reason for this phenomenon is that the BSS approach is applied for crafted instances, and most of the crafted instances are SAT ones.

\section{Conclusions and Future Work}
We defined a new branching strategy selection approach based on vivification ratio of learnt clauses. Through experimental investigation, we analyzed the relationship between different types of instances with the vivification ratio and branching strategy, and found that if the vivification ratio of the instance is very low, then it is more suitable for the instance solved by using more LRB branch strategy. Furthermore, we performed an in-depth empirical analysis that shows that the proposed branching strategy selection approach is robust and allows to solve more instances from recent SAT competitions, especially for main track of 2020 SAT competition. 

In fact, through the analysis of the redundancy of the original clauses for the crafted and application instances, it can be found that the original clauses of the application instances have a higher redundancy ratio. In other words, using vivification method can detect many redundant literals of original clauses during pre-processing for application instances. While generally speaking, the original clauses redundancy ratio of crafted instances are 0. 

Therefore, in future work, the redundancy ratio of original clauses can also be considered as a measure to analyze the types of instances. Specifically, if the redundancy ratio of the original clauses of an instance is 0, the instance is a crafted one, even if the vivification ratio of the learnt clauses of the instance is very high. So for an instance, if the learnt vivification ratio is very high and the original redundancy ratio is also high, we can consider it as an application instance and use more VSIDS for solving it. In summary, the original clauses redundancy ratio and the learnt clause vivification ratio can be used as two parameters to judge the type of instance more accurately, thereby guiding the selection of branching strategies.

\bibliographystyle{alpha}
\bibliography{BSS}
\end{document}